# Optimizing What We Trust: Reliability-Guided QUBO Selection of Multi-Agent Weak Framing Signals for Arabic Sentiment Prediction

**Rabab Alkhalifa**
Department of Computer Engineering
College of Computer Science and Information Technology (CCSIT)
Imam Abdulrahman Bin Faisal University (IAU)
Dammam, Saudi Arabia
raalkhalifa@iau.edu.sa

## Abstract

Framing detection in Arabic social media is difficult due to interpretive ambiguity, cultural grounding, and limited reliable supervision. Existing LLM-based weak supervision methods typically rely on label aggregation, which is brittle when annotations are few and socially dependent. We propose a reliability-aware weak supervision framework that shifts the focus from label fusion to data curation. A small multi-agent LLM pipeline—two framers, a critic, and a discriminator—treats disagreement and reasoning quality as epistemic signals and produces instance-level reliability estimates. These estimates guide a QUBO-based subset selection procedure that enforces frame balance while reducing redundancy. Intrinsic diagnostics and an out-of-domain Arabic sentiment transfer test show that the selected subsets are more reliable and encode non-random, transferable structure, without degrading strong text-only baselines.

## 1 Introduction

Large language models (LLMs) have recently emerged as a powerful source of weak supervision for NLP tasks, enabling the automatic generation of labels, rationales, and confidence estimates at scale (Wei et al., 2022; Wang et al., 2023). This has renewed interest in weak supervision as a practical alternative to costly expert annotation, particularly for tasks where labels are expensive or difficult to define precisely (Frenay and Verleysen, 2014; Ratner et al., 2017; Song et al., 2023). However, many existing weak-supervision frameworks implicitly assume that disagreement among annotators can be resolved through aggregation, often by estimating a single latent "true" label.

This assumption becomes fragile for *socially interpretive* NLP tasks such as framing analysis, stance detection, or political sentiment, where ambiguity and perspective are intrinsic rather than incidental (Pavlick and Kwiatkowski, 2019; Basile et al., 2021). Different annotators—or different prompts applied to the same LLM—may emphasize distinct aspects of the same text, leading to systematic disagreement that reflects competing interpretations rather than annotation error. Collapsing such disagreement into a single label risks discarding valuable information about uncertainty and contestation.

Arabic social media provides a particularly challenging setting. Public discourse around topics such as قيادة المرأة للسيارة ("women driving") intertwines moral, religious, identity-based, legal, and security-oriented arguments. These rhetorical strategies are commonly described in terms of *frames*—structured ways of contextualizing or justifying a position. Framing is closely related to downstream attitudes such as sentiment and stance, making it a useful intermediate representation for modeling social meaning. At the same time, high-quality frame-annotated Arabic datasets remain scarce: annotation guidelines are non-trivial to design, expert labeling is costly, and many instances are genuinely ambiguous.

In this work, we ask a methodological question: *How can LLM-based weak supervision be used to construct more trustworthy training data for framing models, without assuming that all disagreement should be resolved?* Rather than aggregating multiple weak labels into a single probabilistic target, we propose to treat disagreement, confidence asymmetry, and justification quality as *epistemic signals* that inform how much a weak label should be trusted.

We introduce a reliability-aware weak supervision framework built around a small multi-agent LLM pipeline. Two independent LLM framers assign frame labels and provide rationales; a third LLM acts as a critic that evaluates competing explanations and adjudicates a final frame with a rubric-based quality score. From these multi-agent signals, we learn an instance-level reliability esti-mate that reflects the stability and support of each weak label, rather than its assumed correctness.

Having obtained weak labels augmented with reliability estimates, we address a second practical challenge: *which* weakly labeled examples should be used for training. LLM-generated annotation pools are often redundant, imbalanced, and heterogeneous in quality. We therefore cast data curation as a *Quadratic Unconstrained Binary Optimization (**QUBO**)* problem that jointly rewards high-reliability instances, penalizes redundancy via text similarity, and enforces fixed per-frame budgets. Solving this objective yields compact, frame-balanced subsets that are more reliable and less redundant than distribution-matched sampling.

We position our study as a methodological case study rather than a competitive benchmark. All framing labels are synthetically generated by the proposed pipeline, and we do not claim to solve Arabic framing or sentiment at scale. Instead, we evaluate the framework through intrinsic diagnostics and a conservative downstream transfer experiment on a gold-labeled women-driving sentiment dataset. The goal is to test whether reliability-aware selection produces framing signals with non-random, transferable structure, not to outperform strong text-only models.

**Contributions.** Within this scope, our main contributions are:

- a multi-agent LLM weak-supervision pipeline that treats disagreement as epistemic signal rather than noise;
- an instance-level reliability estimation approach derived from multi-agent agreement and justification quality;
- a QUBO-based data selection strategy that integrates reliability, redundancy, and frame balance; and
- an empirical analysis showing that reliability-aware selection yields more stable weak labels and supports downstream transfer without degrading performance.

The remainder of the paper is organized as follows: Section 2 reviews related work; Section 3 presents the multi-agent reliability framework; Section 4 describes QUBO-based subset selection; Section 5 outlines the evaluation protocol; Section 6 details datasets and experimental setup; Section 7 reports results; and Section 8 discusses implications and limitations.

## 2 Related Work

**Weak Supervision and Label Aggregation** Weak supervision replaces costly expert annotation with inexpensive and noisy sources such as heuristic rules, weak classifiers, crowdsourcing, or LLM-based annotators. Classical frameworks such as **Snorkel** (Ratner et al., 2017) model these sources as *labeling functions* (LFs) and estimate their accuracies using a generative label model, often optimized via EM-style procedures (Dawid and Skene, 1979). Subsequent work extends this paradigm by modeling dependencies among LFs (Bach et al., 2017) or by allowing instance-dependent noise characteristics (Cheng et al., 2022).

These approaches generally assume access to many heterogeneous and approximately independent supervision sources whose errors can be statistically disentangled. This assumption is fragile in socially interpretive settings—such as Arabic framing—where labels are subjective, context-dependent, and shaped by cultural and rhetorical perspective. In such regimes, disagreement is common and does not necessarily indicate annotation error.

Our work departs from aggregation-centric weak supervision in two ways. First, we operate in a *small, dependent* multi-agent LLM setting, where each annotator produces not only a label but also a rationale and confidence cues. Second, rather than aggregating annotations into a single probabilistic label, we compute an *instance-level reliability score* that reflects epistemic stability across agents. This reliability signal is not used to infer ground truth, but instead drives a subsequent optimization-based data selection step. In this sense, our framework shifts weak supervision from label aggregation to *selective trust and curation*.

**Reliability, Calibration, and Disagreement** Calibration research examines how well model confidence estimates reflect empirical correctness, showing that modern neural models are often miscalibrated and benefit from post-hoc correction methods (Niculescu-Mizil and Caruana, 2005; Guo et al., 2017). In parallel, annotation modeling studies how label noise arises from annotator expertise, bias, and item difficulty, demonstrating that majority voting fails to capture systematic variation across annotators (Sheng et al., 2008; Paun et al., 2018).

More recent NLP work emphasizes that disagreement can be an informative signal, particularly for subjective or socially grounded tasks. Empirical studies show that disagreement often persists even under repeated annotation, reflecting genuine interpretive ambiguity rather than noise (Pavlick and Kwiatkowski, 2019; Davani et al., 2022). Position papers further argue that collapsing annotations into a single gold label can obscure meaningful variation and lead to misleading evaluation practices (Basile et al., 2021; Uma et al., 2021).

We adopt this epistemic perspective by treating multi-agent LLM annotations as structured evidence rather than conflicting votes. Agreement patterns, confidence asymmetries, and critic feedback are summarized into a scalar reliability estimate per instance. Crucially, reliability in our framework is *not* used to calibrate probabilities or reweight annotators. Instead, it serves as a selection signal within a structured optimization objective, allowing stable and well-supported instances to be prioritized while ambiguous cases are deemphasized.

**Optimization-Based Data Selection and QUBO** Data selection has been widely studied in active learning (Settles, 2009), core-set construction (Sener and Savarese, 2018), dataset pruning (Toneva et al., 2019; Zhang et al., 2023), and diversity-driven subset selection in NLP (Wang et al., 2024). These methods typically aim to improve representativeness or informativeness, but do not incorporate epistemic signals arising from structured multi-agent reasoning.

**QUBO** offers a flexible formalism for jointly encoding item-level utilities and pairwise redundancy penalties, and has been applied to feature and subset selection using classical and quantum-inspired solvers (Nembrini et al., 2021; Aramon et al., 2019; Shaikh et al., 2025). To our knowledge, **QUBO** has not previously been applied to weak supervision data curation under multi-agent LLM annotation.

We adapt **QUBO** to this setting by rewarding instances with higher learned reliability, penalizing redundancy via similarity constraints, and enforcing explicit per-frame selection budgets. Unlike approaches that treat disagreement as a prediction target (Jiang and Marneffe, 2022) or optimize primarily for diversity (Liu et al., 2019; Park et al.), our formulation integrates epistemic reliability directly into a structure-aware selection objective. The resulting subsets are more stable, less redundant, and yield framing-derived representations that transfer effectively to downstream sentiment prediction.

## 3 Reliability-Aware Weak Supervision Framework

We propose a reliability-aware weak supervision framework for framing annotation (Figure 1) that models epistemic uncertainty via multi-agent disagreement and reasoning quality. Instead of collapsing multiple weak annotations into a single label, the framework learns an instance-level estimate of label stability and uses it *only* for data selection (Aroyo and Welty, 2015; Uma et al., 2021; Davani et al., 2022).

The framework outputs a weakly labeled dataset where each instance is associated with (i) an adjudicated frame label and (ii) an instance-level reliability score. Reliability is not used to modify labels or directly reweight training; it is used exclusively to guide subset selection.

The framework has three components: (1) independent multi-agent labeling, (2) critic-based arbitration, and (3) learned reliability estimation.

**Multi-Agent Labeling** Each sentence $x$ is independently annotated by two instruction-tuned LLMs, **Labeler A** and **Labeler B**. Each labeler produces: (i) a frame label from a fixed taxonomy, (ii) a confidence score in $[0, 1]$, and (iii) an evidence-grounded justification.

Formally,

$$\text{Labeler}_A(x) = (\ell_A, c_A, e_A),$$

$$\text{Labeler}_B(x) = (\ell_B, c_B, e_B)$$

where $\ell$ is the predicted frame, $c$ is self-reported confidence, and $e$ is a short evidence span/rationale grounded in the input.

The labelers use different model instances and prompting configurations to encourage partially independent reasoning paths; disagreement is preserved as a potentially informative signal rather than being averaged away.[1]

[1] In our experiments, Labeler A used Qwen-2.5 (3B), Labeler B used Mistral-7B, and the Critic used Gemma-2 (9B), all in instruction-tuned, 4-bit quantized variants.

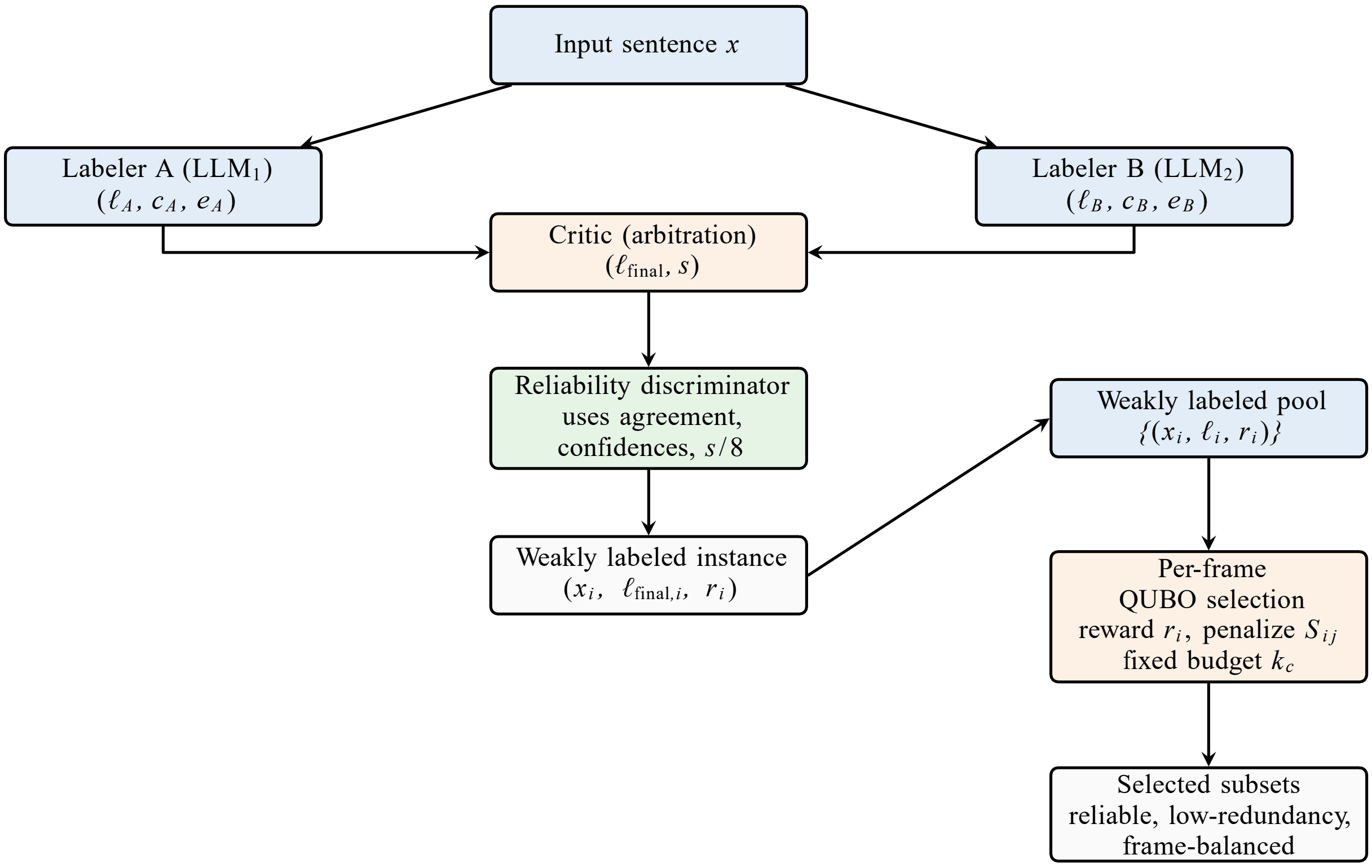

Figure 1: **Reliability-aware weak supervision with QUBO-based data curation.** Two LLM labelers provide labels, confidences, and evidence; a Critic adjudicates and assigns rubric score $s \in \{0, \ldots, 8\}$. A discriminator maps agreement, confidences, and $s$ to reliability $r_i$, which guides per-frame QUBO selection (reward $r_i$, penalize TF–IDF similarity) to yield compact, frame-balanced subsets.

**Critic Arbitration** Disagreement is handled by a third agent, the **Critic**, which adjudicates between competing interpretations by evaluating the supporting evidence and reasoning quality. Rather than voting or averaging labels, the Critic compares the two justifications and selects the frame that is better supported by the text.

Given the labeler outputs, the Critic produces

$$(\ell_{\text{final}}, s) = \text{Critic}(\ell_A, e_A; \ \ell_B, e_B),$$

where $\ell_{\text{final}}$ is the adjudicated frame label and $s \in \{0, \ldots, 8\}$ is a rubric-based reasoning score.

The rubric aggregates four criteria—evidence quality, taxonomy fit, internal coherence, and justification sufficiency—each rated on a three-point scale (0/1/2) and summed to yield a total score $s \in \{0, \ldots, 8\}$. Low scores indicate weak or inconsistent support, while high scores indicate strong epistemic support. This follows work showing that disagreement in semantic annotation often reflects genuine ambiguity or perspective rather than annotator error (Basile et al., 2021; Pavlick and Kwiatkowski, 2019).

**Learned Reliability Estimation** While the Critic resolves disagreement at the instance level, reliability also exhibits recurring global patterns: certain configurations of agreement, confidence asymmetry, and weak justification correlate with unstable labels. To capture these regularities, we train a lightweight **reliability discriminator**.

For each instance $i$, the discriminator uses features derived from the multi-agent process, including: (i) labeler confidences $(c_A, c_B)$, (ii) agreement indicators between Labeler A, Labeler B, and the Critic, (iii) the normalized rubric score $s/8$, and (iv) shallow textual statistics (e.g., sentence length).

A logistic regression model is trained on a pseudo-label that marks instances as *stable* when high-confidence agreement aligns with strong Critic endorsement. The discriminator outputs

$$r_i = P(\text{stable} \mid x_i), \quad (1)$$

reflecting how well-supported the adjudicated label is given the available epistemic evidence. These reliability scores are not used to recalibrate labels; they serve exclusively as selection signals in the QUBO-based subset optimization (Section 4).

## 4 QUBO-Based Subset Selection

The weakly labeled pool is heterogeneous in reliability and contains substantial redundancy (near-duplicates). We therefore curate compact, frame-balanced training subsets using a per-class Quadratic Unconstrained Binary Optimization (QUBO) objective (Figure 1). Selection is performed independently within each frame, enforcing exact frame balance via fixed budgets.

**QUBO objective (per class).** For a frame $c$, let $I_c$ be indices of candidate instances with adjudicated label $c$, and let $z_i \in \{0, 1\}$ indicate whether instance $i \in I_c$ is selected. Each instance has reliability $r_i \in [0, 1]$, and redundancy is measured by TF–IDF cosine similarity $S_{ij}$ computed within the frame. We define the per-class energy

$$E_c(\mathbf{z}) = -\lambda_{\text{rel}} \sum_{i \in I_c} r_i z_i + \lambda_{\text{red}} \sum_{i<j} S_{ij}\, z_i z_j, \quad (2)$$

subject to the fixed-size constraint $\sum_{i \in I_c} z_i = k_c$.

The first term rewards selecting reliable instances, while the second penalizes selecting redundant pairs. For example, if two items have similar $r_i$ but high $S_{ij}$ (near-duplicates), select one; a slightly lower-$r_i$ item may win if less redundant. Solving Eq. (2) independently for each frame enforces exact frame balance by construction.

**Implementation note.** We optimize Eq. (2) using budget-preserving simulated annealing with swap-only local moves: each proposal swaps one selected instance with one unselected instance within the same frame, maintaining $\sum_{i \in I_c} z_i = k_c$ at all times. The energy change is computed from the reliability term and the candidate's interactions under $S_{ij}$ with the current selected set, enabling scalable optimization over large pools.

## 5 Evaluation Protocol

We evaluate our approach as a *methodological contribution* to weak supervision in socially interpretive settings, not as a framing benchmark or a dataset-construction effort. Because all framing labels in the synthetic corpus are produced by a multi-agent LLM pipeline, predictive scores on that corpus should be interpreted as *internal consistency* with the generating weak signals rather than semantic correctness.

We evaluate in two complementary settings.

| Dataset | Size | Years | Task |
|---|---|---|---|
| Weak Framing (Synthetic) | 2,733 | 2015–2019 | Framing |
| Women-Driving (Gold) | 2,442 | 2012–2017 | Sentiment |

Table 1: Datasets used in this study. Detailed label distributions are provided in Appendix A.

**Intrinsic diagnostic evaluation (subset quality).** We quantify how reliability-aware QUBO selection changes the properties of the selected training signal under *synthetic* framing supervision. Concretely, we train a lightweight TF–IDF + logistic regression framing classifier on either (i) a QUBO-selected subset or (ii) a size-matched *distribution-matching* baseline, and evaluate against the generating weak labels to obtain a diagnostic Macro-F1. To characterize redundancy, we also report mean pairwise TF–IDF cosine similarity within the selected subset (Section 7.3, Figure 4).

**Out-of-domain downstream evaluation (conservative transfer).** We test whether QUBO-curated *synthetic* framing signals encode *transferable structure* on a human-labeled task. Using the gold women-driving sentiment dataset, we represent each tweet with BoW text features and a frame-probability vector produced by a framing model trained on either QUBO-selected data or the size-matched baseline. We then train sentiment classifiers (logistic regression) under seven configurations: text only (S0), text + DistMatch/QUBO framing features (SD, SQ), negative controls with noise or shufled QUBO features (SN, SQshuf), and framing-only models (FD, FQ). Results are reported on the held-out gold test split in Section 7.4 (Table 3).

Overall, the downstream goal is not to outperform strong text-only baselines, but to verify that QUBO-selected synthetic framing features (i) do not systematically degrade performance and (ii) outperform noise and shufling controls, consistent with non-random structure.

## 6 Experimental Setup

This section describes the datasets and experimental settings used across the intrinsic diagnostic study and the out-of-domain transfer evaluation.

### 6.1 Datasets

We use two datasets: a synthetic weakly labeled Arabic framing corpus and a human-annotated women-driving sentiment dataset.

| Group | $n$ | Mean $r_i$ | Mean critic |
|---|---|---|---|
| High reliability | 1,360 | 0.99 | 6.32 |
| Low reliability | 1,373 | 0.01 | 5.10 |

Table 2: **Learned reliability groups.** High-reliability examples cluster near $r_i \approx 1$ with higher critic rubric scores; low-reliability examples cluster near $r_i \approx 0$ with lower scores.

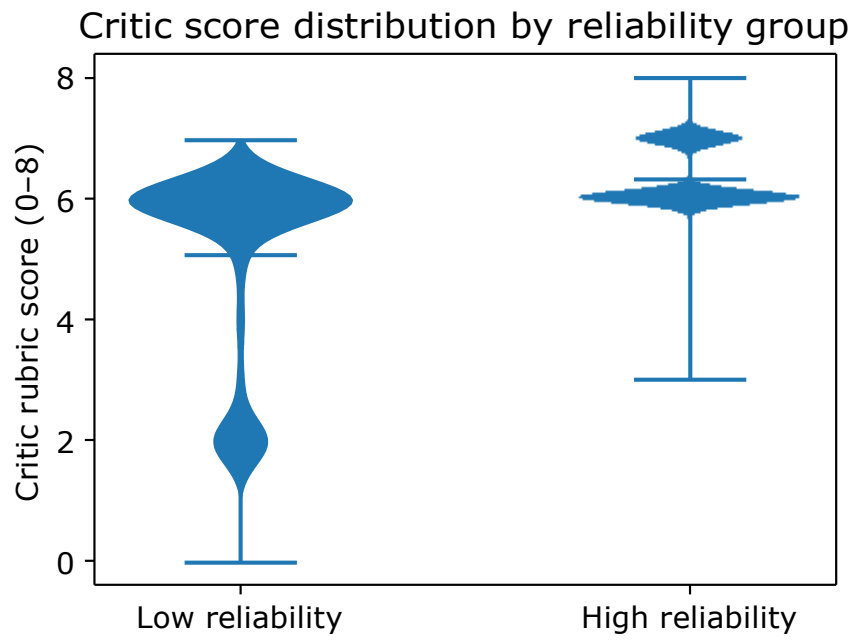

Figure 2: **Critic score distributions (0–8) by reliability group.** High-reliability instances concentrate at higher rubric scores; low-reliability instances are broader and centered lower.

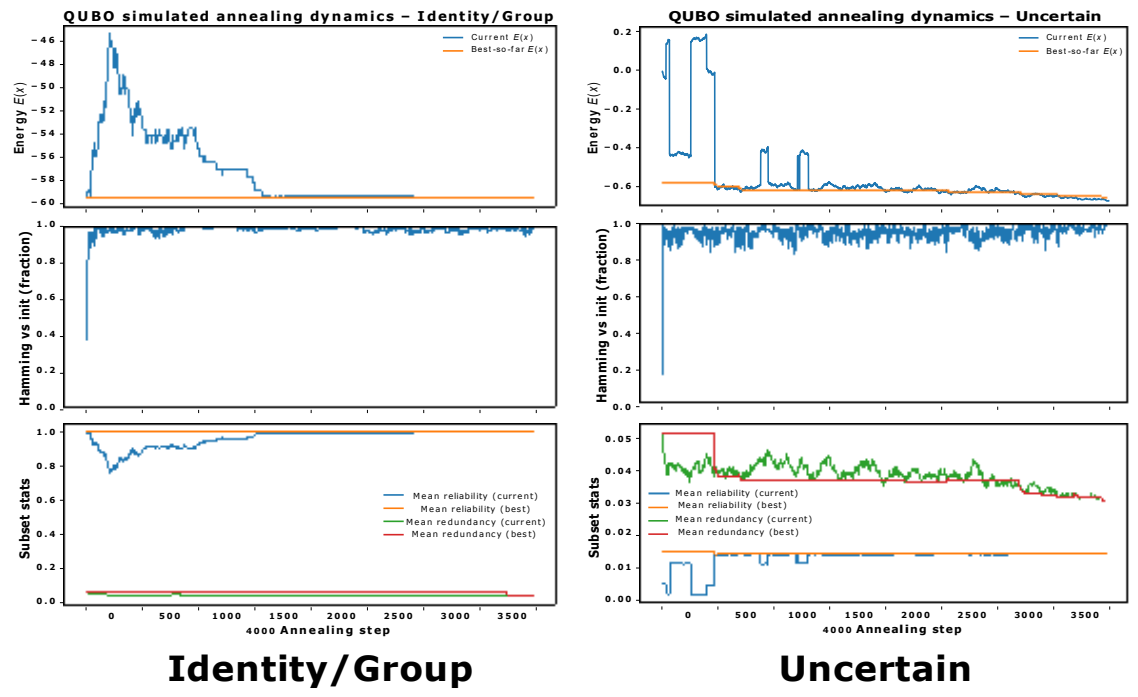

Figure 3: **Simulated annealing dynamics across two frames. Left:** *Identity/Group* (high-resource) transitions from early exploration to stable convergence, largely replacing the warm start while increasing reliability and reducing redundancy. **Right:** *Uncertain* (mid-sized, noisy) shows stronger energy oscillations from competing minima, yet converges to a coherent low-energy subset with higher reliability and reduced redundancy.

**Weak Framing (Synthetic).** We construct a synthetic temporal framing corpus by prompting an LLM to generate short, aspect-focused Arabic statements conditioned on dominant sociopolitical themes extracted from a longitudinal Twitter collection (2015–2019). After deduplication, the dataset contains 2,733 instances annotated using our multi-agent framework. The resulting label distribution is highly imbalanced, motivating the use of reliability-aware and diversity-constrained subset selection.

**Women-Driving Sentiment (Gold).** For out-of-domain evaluation, we use a human-annotated women-driving sentiment dataset (Addawood et al., 2018) containing 2,442 tweets from 2012–2017 with positive, neutral, and negative labels. This dataset is not weakly supervised and is used solely to assess the transferability of framing representations learned from synthetic data. Both datasets are split into 80/20 train/test partitions using stratified sampling.

Further details about the datasets are provided in the appendix A.

# 7 Results

## 7.1 Multi-agent Framework

We analyze weak labels produced by the two labelers, the Critic, and the learned reliability discriminator. After surface-form de-duplication, the corpus contains 2,733 unique examples, each with a final frame label, calibrated confidence, and reliability probability $r_i \in [0, 1]$.

For interpretability, we form two groups using the discriminator: *low reliability* ($r_i < 0.33$) and *high reliability* ($r_i \geq 0.66$). The split is nearly even (1,373 vs. 1,360), but the profiles differ substantially (Table 2). High-reliability examples have $r_i$ near 1 (mean 0.99) and higher critic scores (mean 6.32), while low-reliability examples have $r_i$ near 0 (mean 0.01) with lower critic scores (mean 5.10). This alignment indicates that $r_i$ tracks the Critic's rubric assessments rather than simply mirroring confidence.

Figure 2 shows the critic score distributions. High-reliability examples concentrate near the upper range (roughly 6–8), whereas low-reliability examples are more dispersed and centered lower, with limited overlap. These results support using $r_i$ as a selection signal for QUBO curation.

## 7.2 QUBO Optimization Dynamics Across Frames

We examine simulated annealing trajectories for two representative frames to illustrate QUBO behavior under different regimes: a high-resource frame (*Identity/Group*) and a mid-sized ambiguous frame (*Uncertain*) (Figure 3).

**Identity/Group.** The sampler shows smooth annealing: after brief exploratory fluctuations, en-ergy decreases steadily. The Hamming curve approaches 1.0, indicating that most warm-start items are replaced. Reliability rises toward 1.0 and redundancy falls, consistent with effective selection in a well-structured, data-rich frame.

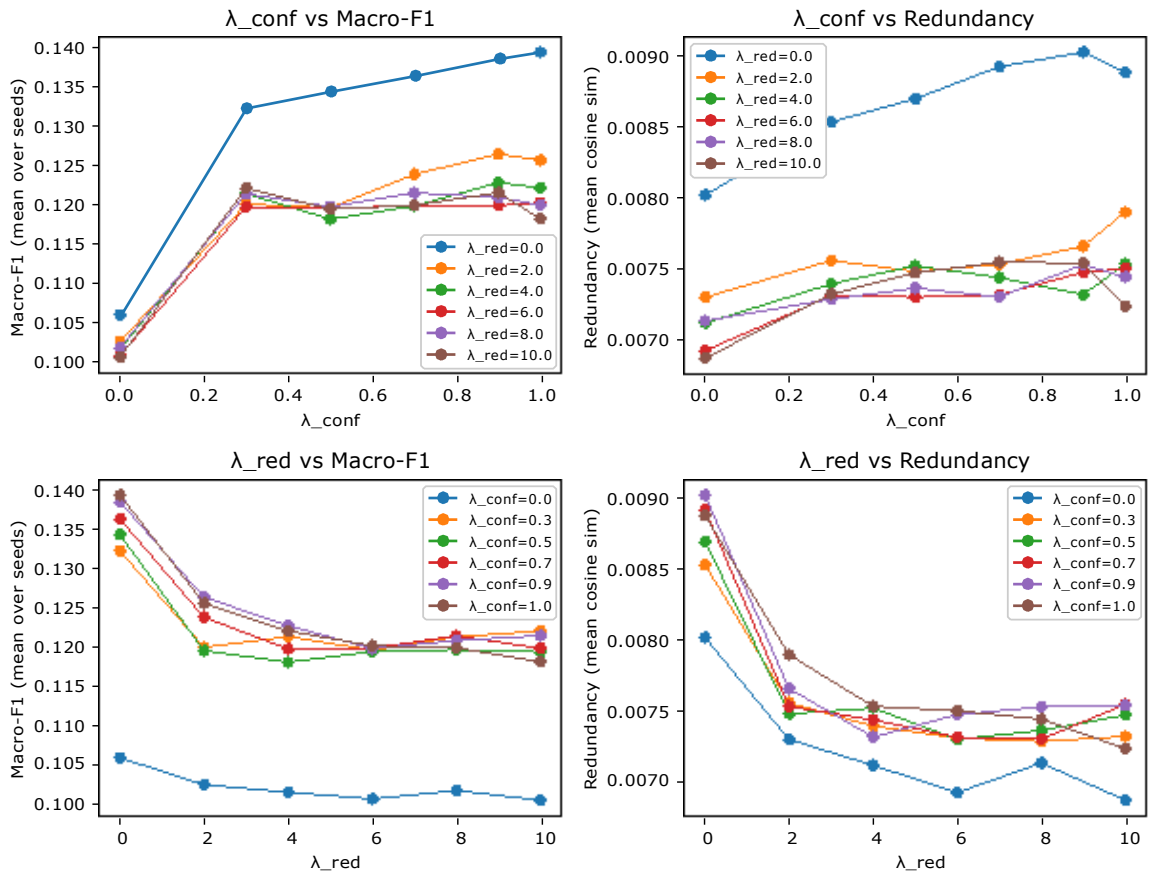

Figure 4: **QUBO hyperparameter trade-offs.** Top: effect of $\lambda_{\text{conf}}$ on Macro-F1 and redundancy across multiple $\lambda_{\text{red}}$ settings. Bottom: effect of $\lambda_{\text{red}}$ across different $\lambda_{\text{conf}}$ regimes. Both parameters exhibit mid-range values that consistently improve diagnostic performance while suppressing redundancy.

**Uncertain.** Energy oscillates early, reflecting many competing local minima from label ambiguity and noisy reliability signals. Despite the irregular landscape, the sampler converges to a stable low-energy subset that again largely replaces the warm start and yields higher mean reliability with lower redundancy.

**Low-resource frames.** When $k \geq n_{\text{frame}}$ (e.g., *Public Health/Safety*, *Economic/Cost–Benefit*, *Security/Threat*), the feasible set collapses to a single solution and trajectories are flat; we omit these boundary cases.

### 7.3 QUBO Trade-Offs

We performed a systematic sweep over QUBO hyperparameters to study how the reliability weight $\lambda_{\text{conf}}$ and redundancy penalty $\lambda_{\text{red}}$ shape the objective and the quality of selected weak labels. For each setting, we measure intrinsic diagnostic Macro-F1 and mean pairwise cosine similarity of the selected subset (redundancy).

**Effect of $\lambda_{\text{conf}}$.** The top row of Figure 4 shows that when $\lambda_{\text{conf}}$ is small, the optimiser under-uses the reliability signal and behaves like distribution matching, yielding lower Macro-F1 and higher redundancy. As $\lambda_{\text{conf}}$ increases, Macro-F1 improves and stabilises, indicating that agreement-based reliability provides consistent guidance. Very large values bias the optimiser toward repeatedly selecting a small set of highly prototypical sentences, raising redundancy.

| Method | Accuracy | Macro-F1 |
|---|---|---|
| S0 (text only) | 0.6319 | 0.6237 |
| SD (text + DistMatch) | 0.6278 | 0.6193 |
| SN (text + noise) | 0.6094 | 0.6039 |
| SQshuf (text + shufled QUBO) | 0.6237 | 0.6161 |
| **SQ (text + QUBO)** | **0.6339** | **0.6254** |
| FD (frames only, DistMatch) | 0.4049 | 0.3989 |
| **FQ (frames only, QUBO)** | **0.4397** | **0.4177** |

Table 3: **Downstream sentiment classification under multiple feature configurations**. Results on the held-out gold test split using a shared BoW logistic regression classifier.[2]

**Effect of $\lambda_{\text{red}}$.** The bottom row of Figure 4 shows that with $\lambda_{\text{red}} \approx 0$, the optimiser selects near-duplicates, especially in high-confidence regimes. Increasing $\lambda_{\text{red}}$ strongly suppresses redundancy while largely preserving Macro-F1, revealing a broad operating region. Excessively large penalties eventually degrade performance by forcing selection toward lower-quality or borderline examples.

Additional visualizations (Pareto frontier and $\Delta$F1 advantage map) are provided in Appendix B.

### 7.4 Downstream influence of QUBO-selected framing features

We test whether QUBO-selected *synthetic* framing signals provide useful auxiliary information for a supervised downstream task. Using the gold women-driving sentiment dataset, we compare seven feature configurations: text-only (S0), text + DistMatch/QUBO framing features (SD, SQ), two negative controls (SN, SQshuf), and framing-only models (FD, FQ). All systems use the same BoW logistic regression backbone and hyperparameters.

Table 3 reports accuracy and macro-F1 on the held-out test split. The text-only baseline is strong (S0 macro-F1 = 0.624). Adding framing features yields comparable performance: SQ is slightly higher than S0 and SD, but we do not claim a statistically significant improvement over text.

Crucially, SQ outperforms both negative controls. Injecting Gaussian noise (SN) reduces

[2]Framing labels are fully synthetic (LLM-generated). This downstream experiment is a conservative stress test of feature transfer rather than a task-optimized benchmark.

macro-F1 to 0.604, and shufling QUBO frame probabilities (SQshuf) reduces macro-F1 to 0.616, yielding the ordering SQ > SQshuf > SN. This pattern indicates that QUBO-selected framing vectors encode *non-random, aligned structure*, even if the effect is modest in a BoW setting.

Framing-only models further isolate this signal. While overall performance is lower than text-based systems, both exceed chance, and the QUBO variant (FQ) consistently outperforms the DistMatch baseline (FD), suggesting that QUBO produces more informative framing representations when lexical cues are removed.

Overall, QUBO-selected synthetic framing features provide a small but systematic downstream signal: they are robust to noise, sensitive to shufling, and stronger than distribution matching in framing-only settings.

## 8 Discussion

Our experiments support an optimization-first view of weak supervision for socially interpretive tasks: multi-agent LLM annotation yields usable *epistemic metadata*, and QUBO subset selection converts these signals into compact, frame-balanced subsets with reduced redundancy.

**Epistemic metadata from multi-agent supervision.** The discriminator partitions the synthetic pool into two regimes. High-reliability instances cluster near $r_i \approx 1$ and receive stronger critic rubric scores, while low-reliability instances cluster near $r_i \approx 0$ with weaker critic assessments (Table 2; Figure 2). This aligns with prior work that interprets persistent disagreement in subjective NLP as ambiguity or perspective rather than simple error (Pavlick and Kwiatkowski, 2019; Davani et al., 2022). We do not treat weak labels as semantically correct; reliability is a *selective-trust* signal for curation, not a proxy for gold accuracy.

**Why QUBO selection improves subset quality.** Redundancy is pairwise and is therefore poorly controlled by pointwise heuristics or distribution matching. Hyperparameter sweeps show that increasing $\lambda_{\text{conf}}$ improves intrinsic diagnostic agreement, while $\lambda_{\text{red}}$ suppresses near-duplicates with limited Macro-F1 loss across a broad operating region (Figure 4).

**Conservative transfer beyond framing.** On the gold women-driving sentiment task, QUBO-derived framing features remain competitive with the text-only baseline and outperform noise and shufling controls; framing-only models also benefit from QUBO selection (Table 3). Because framing supervision is synthetic and the setup is conservative, we interpret this as evidence of *non-random* transferable structure, not improved framing accuracy.

**Relation to classical weak supervision.** Label-model-centric frameworks typically rely on many heterogeneous sources whose accuracies and dependencies can be estimated. Here, a small set of adaptive, prompt-driven LLM annotators makes explicit dependency modeling brittle. We therefore shift emphasis from aggregation to *curation*: compute instance-level reliability and use it to drive fixed-budget, frame-balanced selection, yielding cleaner training subsets without an explicit dependency graph.

**Limitations and availability.** Our QUBO objective scales quadratically with the number of candidates, and our empirical evidence is currently limited to LLM-generated synthetic framing labels and a single downstream transfer setting. Future work should explore approximate and/or decomposable solvers to improve scalability, run broader stress tests across agent configurations, model choices, and prompt variants, and incorporate light-weight human calibration when semantic validity is critical. To support reproducibility, we release the datasets, model versions, and prompts used in our experiments.[3]

## 9 Conclusion

We introduced a reliability-aware weak supervision framework that pairs multi-agent LLM annotation with a QUBO-based subset selector. By treating agreement, critic rubrics, and rationale consistency as epistemic evidence, the selector curates fixed-budget subsets that are more reliable and less redundant than a size-matched distributional baseline. A conservative out-of-domain transfer test on gold-labeled women-driving sentiment indicates that framing features learned from QUBO-selected data encode non-random structure, outperforming noise and shufling controls without degrading strong text-only baselines. Future work will focus on scaling QUBO selection and incorporating light human calibration.

[3] https://github.com/Rababalkhalifa/OptimizingWhatWeTrust

# A Dataset Construction and Statistics

This appendix provides additional details on dataset construction, label distributions, and data splits, omitted from the main paper for space reasons.

## A.1 Synthetic Weak Framing Dataset

**Data Source and Generation.** We construct a synthetic Arabic framing corpus conditioned on temporal public discourse. Starting from a longitudinal Arabic Twitter collection spanning 2015–2019, we extract dominant socio-political themes per year and prompt an LLM to generate short, aspect-focused statements reflecting these themes. The generation process avoids the reuse of user-authored content and is intended to capture realistic framing patterns rather than reproduce original tweets.

After deduplication, the resulting corpus contains 2,733 sentences.

**Framing Taxonomy Discovery.** To identify a stable framing taxonomy, we sample approximately 80 sentences from the synthetic corpus and annotate them using the proposed multi-agent framework. The agents consistently converged on seven framing categories, which are fixed and enforced throughout the full weak supervision pipeline.

**Label Distribution.** Applying the multi-agent framework to the full corpus yields a highly imbalanced label distribution, summarized in Table 4. The distribution is dominated by *Identity/Group* and *Moral/Religious* frames, with several minority categories below 4%. This imbalance motivates the need for frame-balanced subset selection.

| Frame | Proportion (%) |
|---|---|
| Identity / Group | 50.2 |
| Moral / Religious | 29.0 |
| Uncertain | 11.5 |
| Public Health / Safety | 3.3 |
| Rights / Justice | 2.9 |
| Economic / Cost–Benefit | 2.5 |
| Security / Threat | 0.6 |

Table 4: Label distribution of the synthetic weak framing dataset.

**Train/Test Split.** We perform an 80/20 stratified train/test split, preserving both temporal and label proportions. The split statistics are reported in Table 5.

| Dataset / Split | Instances | Percent |
|---|---|---|
| Weak Framing – Train | 2,186 | 79.9% |
| Weak Framing – Test | 547 | 20.1% |

Table 5: Train/test split for the synthetic weak framing dataset.

## A.2 Women-Driving Sentiment Dataset (Gold)

For out-of-domain evaluation, we use the women-driving sentiment dataset introduced by Addawood et al. (2018). The dataset contains 2,442 Arabic

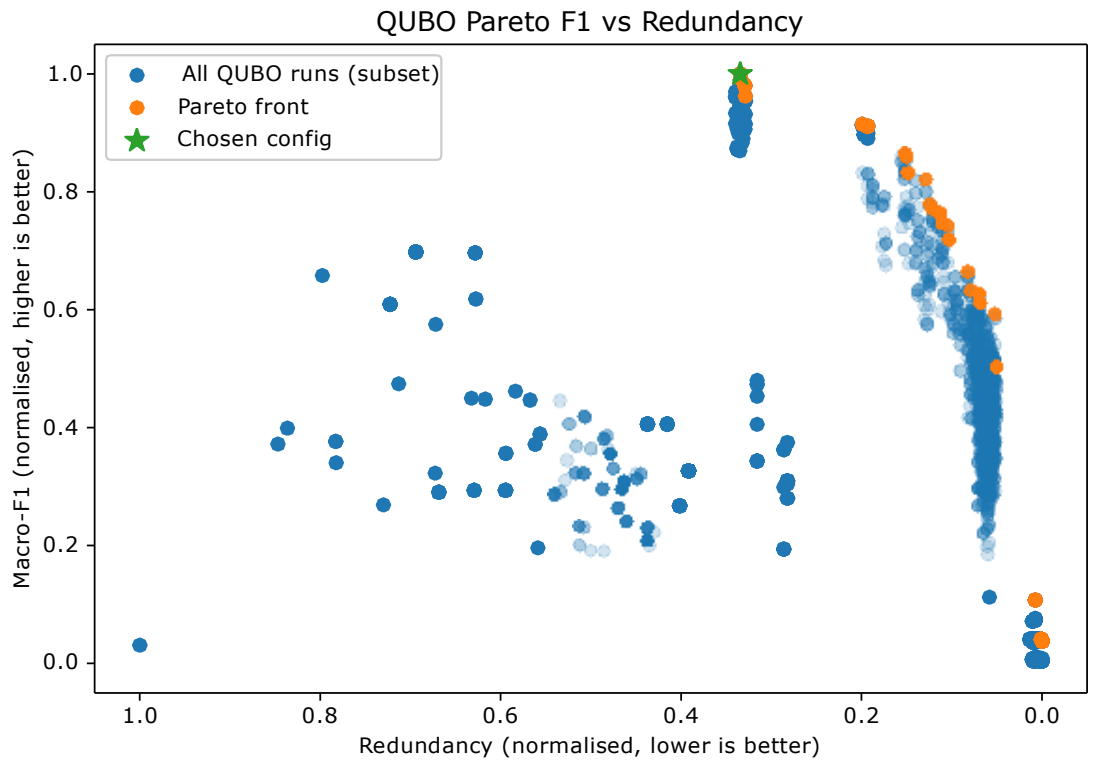

Figure 5: **Pareto frontier of QUBO configurations.** Each point corresponds to a QUBO setting plotted by Macro-F1 (higher is better) and redundancy (lower is better). Highlighted points denote Pareto-efficient, non-dominated solutions. The selected configuration (star) achieves the highest Macro-F1 among all Pareto-efficient settings, representing an accuracy-focused choice within the non-dominated region.

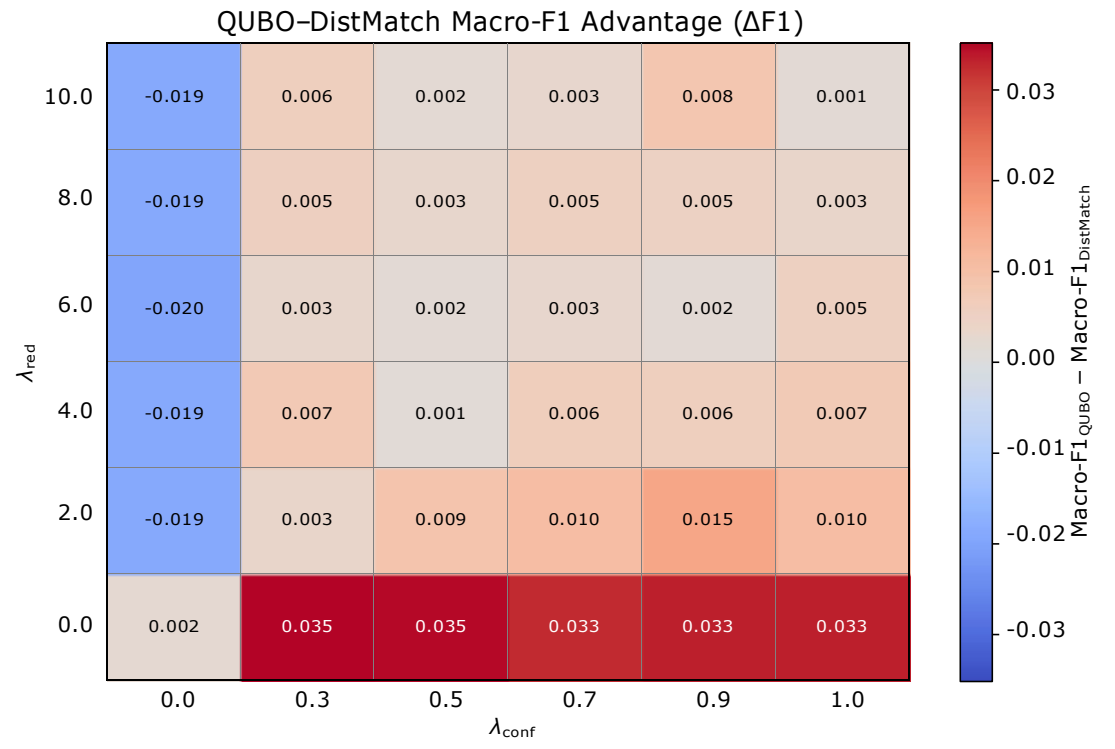

Figure 6: **QUBO vs. DistMatch: $\Delta$F1 advantage map.** Each cell shows $\Delta$F1 on diagnostic Macro-F1 for matched ($\lambda_{\text{conf}}, \lambda_{\text{red}}$) settings (warm = QUBO higher Macro-F1; cool = DistMatch higher Macro-F1).

tweets spanning 2012–2017 with three sentiment labels: positive, neutral, and negative.

After removing duplicates, the dataset contains 2,442 unique tweets. The label distribution is: 1,002 positive (41.0%), 912 neutral (37.4%), and 528 negative (21.6%). We then perform an 80/20 stratified train/test split by sentiment label, yielding 1,953 training and 489 test instances. This dataset is not weakly supervised and is used solely to assess whether framing representations learned from synthetic data encode transferable structure.

# B Additional QUBO Diagnostics

## B.1 Pareto frontier of QUBO configurations

**Pareto frontier.** To identify principled operating points under competing objectives, we visualised all QUBO configurations in the accuracy–redundancy plane and computed the Pareto frontier of non-dominated solutions (Figure 5). A configuration is Pareto-efficient if no alternative simultaneously achieves higher Macro-F1 and lower redundancy. The resulting frontier forms an upper-left boundary of the configuration space, reflecting the inherent trade-off between predictive performance and redundancy. Among these Pareto-efficient configurations, we select the setting with the highest Macro-F1 (starred), prioritising accuracy while ensuring that redundancy is not dominated. This single operating point is used consistently across downstream experiments to avoid frame-specific tuning and to preserve experimental comparability.

## B.2 QUBO vs. DistMatch: $\Delta$F1 advantage map

We compared QUBO against the distribution-matching baseline by computing the diagnostic Macro-F1 difference for each matched ($\lambda_{\text{conf}}, \lambda_{\text{red}}$) setting:

$$\Delta\text{F1} = \text{Macro-F1}_{\text{QUBO}} - \text{Macro-F1}_{\text{DistMatch}}.$$

Figure 6 shows the resulting advantage map (warm = QUBO better, cool = DistMatch better). Across most of the grid, QUBO yields a small but consistently positive advantage (typically $\Delta\text{F1} \approx 0.001$–$0.015$) for $\lambda_{\text{conf}} \geq 0.3$, while the corner case $\lambda_{\text{conf}} = 0$ is uniformly negative for $\lambda_{\text{red}} > 0$. Notably, the strongest gains occur at $\lambda_{\text{red}} = 0$ with mid-to-high $\lambda_{\text{conf}}$ (peaking around $\Delta\text{F1} \approx 0.033$–$0.035$), whereas increasing $\lambda_{\text{red}}$ reduces the magnitude of the advantage but keeps it positive over a broad region.

Overall, the heatmap indicates that incorporating reliability weighting (nonzero $\lambda_{\text{conf}}$) provides robust improvements over DistMatch, and that moderate redundancy penalties trade off some of that gain for lower redundancy, consistent with the trade-off analysis.